\documentclass[letterpaper, 10 pt, conference]{ieeeconf}  
\usepackage{graphicx}
\IEEEoverridecommandlockouts                              
\title{\LARGE \bf Surface Type Estimation from GPS Tracked Bicycle Activities
}

\author{Nitish Nag$^{1,3}$, Vaibhav Pandey$^{1}$, Aishwarya Manjunath $^{2}$, Avinash Vaka$^{2}$, Ramesh Jain$^{1}$
\thanks{*This work was not supported by any organization}
\thanks{$^{1}$ Donald Bren School of Information and Computer Science,
        University of California, Irvine, United States of America
        {\tt\small vaibhap1 at uci.edu}}%
\thanks{$^{2}$ Department of Computer Science, PES University,
        Bangalore, India
        {\tt\small aishwarya.9711 at gmail.com, vavinash0497 at gmail.com}}%
 \thanks{$^{3}$ Medical Scientist Training Program, School of Medicine, University of California, Irvine, United States of America,
        {\tt\small nagn at uci.edu}}%
}

\begin{document}

\maketitle
\thispagestyle{empty}
\pagestyle{empty}

\begin{abstract}
Road conditions affect both machine and human powered modes of transportation. In the case of human powered transportation, poor road conditions increase the work for the individual to travel. Previous estimates for these parameters have used computationally expensive analysis of satellite images. In this work, we use a computationally inexpensive and simple method by using only GPS data from a human powered cyclist. By estimating if the road taken by the user has high or low variations in their directional vector, we classify if the user is on a paved road or on an unpaved trail. In order to do this, three methods were adopted, changes in frequency of the direction of slope in a given path segment, fitting segments of the path, and finding the first derivative and the number of points of zero crossings of each segment. Machine learning models such as support vector machines, K-nearest neighbors, and decision trees were used for the classification of the path. We show in our methods, the decision trees performed the best with an accuracy of 86\%. Estimation of the type of surface can be used for many applications such as understanding rolling resistance for power estimation estimation or building exercise recommendation systems by user profiling as described in detail in the paper. 
\end{abstract}

\section{INTRODUCTION}
Obtaining terrain data is usually done through satellite images and further analysis of these images. Unfortunately, the computing requirements to do this at scale are not feasible due to the heavy computing needs of image processing. A bicycle and a mobile phone are the most commonly owned transportation and electronic devices in the world, including people who live in remote places\cite{c25}\cite{c26}. The data collected from a mobile phone GPS can be used for obtaining useful information about the surrounding terrain features and type of surface. Determining the type of surface can be useful for many applications such as estimating the rolling resistance and friction coefficient, which is otherwise difficult and complex to estimate. These road features can also be included in mapping applications so that users can understand road conditions before travel.

In this paper, we have explored how the GPS coordinates (including altitude) data of the path taken by a bicycle rider can be sufficient for classifying the type of surface between paved and dirt roads. This can be done by using simple concepts in an inexpensive way unlike the image processing techniques applied onto the satellite images. Further classification of the path has been done using supervised machine learning techniques.

\section{RELATED WORK}
Computer vision provides us with solutions to various image understanding tasks such as estimating urban/rural areas, estimation of population density and finding land cover. According to \cite{c15}, over 6 million images tagged with geo location were considered and a data driven scene matching approach was adopted. An automated method to determine terrain model has been proposed in \cite{c1} to best approximate the true data points of the surface models.

Data from a mountain rangeland with Landsat data and 251 sampling sites from central Argentina was considered for surface analysis \cite{c16}. Here, 8 land cover units were defined in terms of spectral information and also ecologically meaningful units in terms of structural types was considered. Classification was done using discriminant functions and maximum likelihood functions to compare against field validations. Ground survey methods such as electronic tachymetry, GPS and terrestrial laser scanning have been used for terrain modeling as well \cite{c17}.

For identification of a road in a given image, Gabor filters have been used to obtain texture orientation for every pixel and edge detection with a vanishing point constraint has been used for finding boundaries of the road in \cite{c18}. Here, 1003 images were used for identifying road regions. Road area detection algorithm from color images has been proposed in \cite{c19} and it involves edge detection using the intensity of the image and analysis of the full color image to determine the road area.

Estimating the nature of road surface becomes an important factor in improving the road infrastructure. An emphasis of use of real time data from the smart phone in estimating the nature of road surface by detecting potholes on roads \cite{c20}. Mobile phones have been used to collect data such as spectral and temporal features of the road along with the vehicle speed to classify road anomalies using support vector classifiers \cite{c21}.

Wavelet analysis has been used for road profiling and in identifying roughness of road surface, detecting potholes and cracks \cite{c22}. Evaluation of road quality has also been demonstrated through GPS data and accelerometer data collected from anonymous drivers \cite{c23}.

In this paper we have experimented with using only GPS data and elevation data from cyclists in evaluating the type of surface of the path. We test three classification techniques: support vector machines, K nearest neighbors, and decision trees. Our aim is to distinguish a dirt/mud path from a developed paved road. 

\section{\textbf{METHODOLOGY}}

The flowchart Fig.1 shows the methodology taken. For each model we use the below classification methods.

\begin{itemize}
\item Support Vector Machines \cite{c3}
\item K nearest neighbors \cite{c4}
\item Decision Trees \cite{c5}
\end{itemize}
\begin{figure}[!ht]
\centering
\textbf{\includegraphics[width=\linewidth,height=9cm]{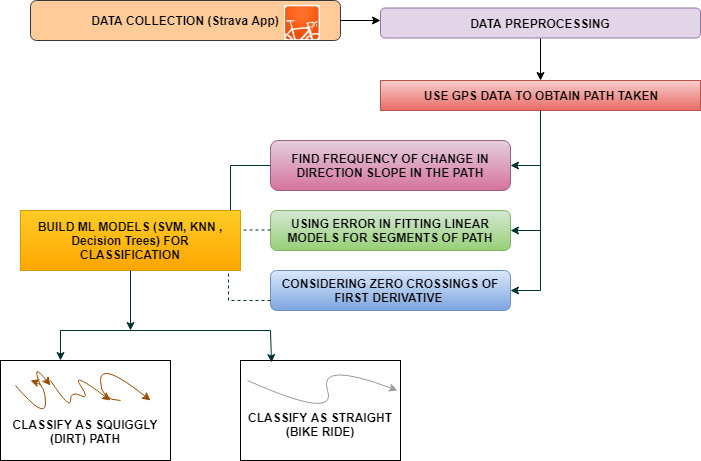}
\caption{\label{Figure 1} Flowchart}}
\end{figure}
\subsection{\textbf{Data Collection}} Strava is the main source of activities. The app uses GPS to track all the physical activities like cycling, running, and swimming. The app users can record their activities and view their statistics. Data from 44 athletes is collected after authorization. This data includes personal information like age , weight , height , waist circumference. Activity data streams are collected using the Strava application. The streams include power, cadence, temperature , altitude , GPS coordinates during the ride, velocity and distance \cite{c2}. From this, we use only GPS and altitude to classify the road.

\subsection{\textbf{Data Preprocessing}} The data collected includes all the activities recorded by the user. From this, only bike rides are selected for estimating the road surface. Currently, 115 rides were selected and these rides were manually classified as squiggly and straight to obtain the ground truth. Among these, 50 \% were considered for training and the rest for testing.

\subsection{Method}
\begin{itemize}

\item \textbf{Assumptions}: The path taken by a bicycle rider on a dirt path is more non-uniform and more squiggly than a bike road which is relatively more uniform and straight. The same was observed from the maps on Golden Cheetah\cite{c24}  for the data collected through Strava App. Figure 2 shows a dirt path and a straight path. This was plotted using the X-Y coordinates from the GPS. It can be seen from Figure 2 that the dirt path taken by the rider is more squiggly.
\newline
\end{itemize}

\begin{figure}
  \textbf{\includegraphics[width=\linewidth]{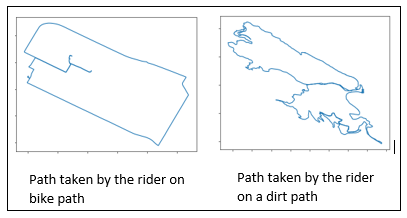}
  \caption{Paths.}}
\end{figure}
To identify if a path is squiggly or straight, the following methods are adopted:
\newline
\subsubsection{\textbf{Using frequency of change in direction of slope}}
\begin{itemize}
\item The path taken by the rider is divided into segments. The length of each segment is 1 percent of the total distance taken by the rider.
\item In each of these segments, the points where slope increases or decreases is noted. This way we are able to get the peaks in the path taken.
\item The variation in the direction of path is more for a squiggly path than a straight path. The frequency of change in slopes is considered. Figure 3 and 4 show the variation in direction of slope of a squiggly and straight path respectively.
\begin{figure}
  \textbf{\includegraphics[width=\linewidth]{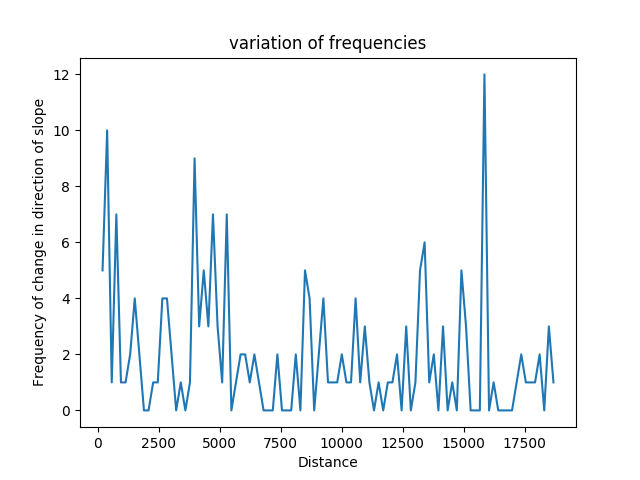}
  \caption{Frequency of variation in direction of slope in segments in a squiggly path}}
\end{figure}
\begin{figure}
  \textbf{\includegraphics[width=\linewidth]{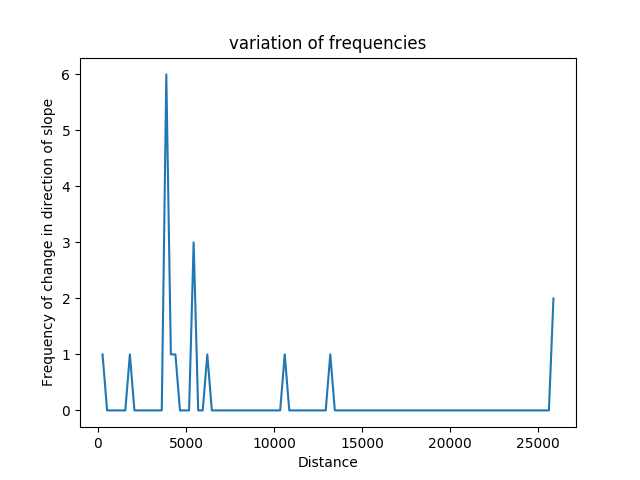}
  \ caption{Frequency of variation in direction of slope in segments in a straight path}}
\end{figure}
\item A threshold for the frequency in change of direction of slope needs to be set to distinguish a squiggly path from a straight path. But, setting the threshold manually can be tedious task. Instead, we have adopted the usage of supervised classification techniques.

For each of the three models, table 1 shows the summary statistics when the ratio of train to test is 50:50
\begin{table}
\begin{center}
 \begin{tabular}{||c c c c||} 
 \hline
 Model & Accuracy & Precision & Recall\\ [0.1ex] 
 \hline\hline
 SVM &   80.70\% & 81\% & 81\%\\ 
 \hline
 KNN with K=3 &  69\% & 76\% & 68\%\\
 \hline
 Decision Trees & 74\% & 79\% & 74\%\\
 \hline
\end{tabular}
\textbf{\caption{Classification Report for method 1}}
\end{center}
\end{table}
\end{itemize}

Comparison of the models is shown in the figure 5. ROC curve analysis is considered for evaluating the classifiers \cite{c14}. From figure 5 it can be seen that the area under the ROC curve for SVM model was the highest. Therefore SVM performed the best compared to all other models.
\begin{figure}
  \textbf{\includegraphics[width=\linewidth]{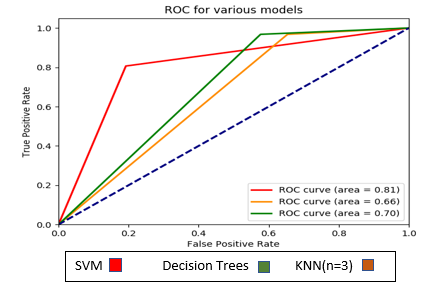}
  \caption{ROC Curve}}
\end{figure}
\newline
\subsubsection{\textbf{Using error in fitting the curve with Linear models for the segments}}
\begin{itemize}
\item The path taken by the rider is divided into segments. The length of each segment is 1 percent of the total distance taken by the rider.
\item For each of these segments, a linear regression model \cite{c6} built.
\item In each segment 50 percent of the data is considered for training and the rest for testing.
\item The root mean square error obtained by testing the regression model is considered. A squiggly segment being more non-uniform has more root mean square error than  the straight path. Figure 6 and 7 show a linear regression model being fitted in a given segment of a squiggly and straight path respectively. As it can be seen, the line fits the actual data points very well for a straight path as compared to a squiggly path.
\newline
\begin{figure}
  \textbf{\includegraphics[width=\linewidth]{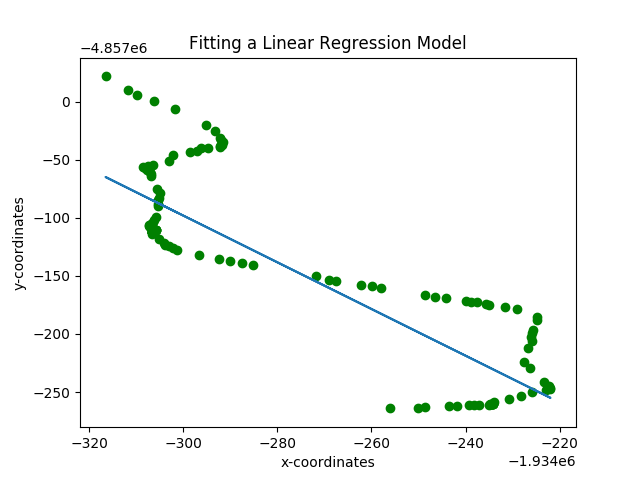}
  \caption{Fitting Linear model in a segment of a squiggly path}}
\end{figure}
\begin{figure}
  \textbf{\includegraphics[width=\linewidth]{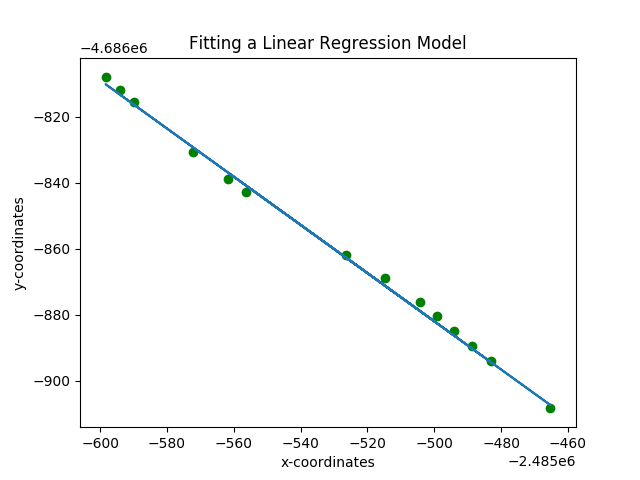}
  \caption{Fitting Linear model in a segment of a straight path}}
\end{figure}
\end{itemize}
For each of these models, the Table 2 shows the summary statistics when the ratio of train to test is 50:50
\newline

Comparison of the models is show in the figure 8. From this figure it can be seen that the area under the ROC curve for Decision Trees model was the highest. Therefore, for this method, decision trees performed the best compared to all other models.
\begin{table}
\begin{center}
 \begin{tabular}{||c c c c||} 
 \hline
 Model & Accuracy & Precision & Recall\\ [0.4ex] 
 \hline\hline
 SVM &   54\% & 29\% & 54\%\\ 
 \hline
 KNN with K=3 &  86.54\% & 87\% & 87\%\\
 \hline
 Decision Trees & 86.53\% & 87\% & 87\%\\
 \hline
\end{tabular}
\textbf{\caption{ Classification Report using method 2}}
\end{center}
\end{table}
\newline
\begin{figure}
 \textbf{\includegraphics[width=\linewidth]{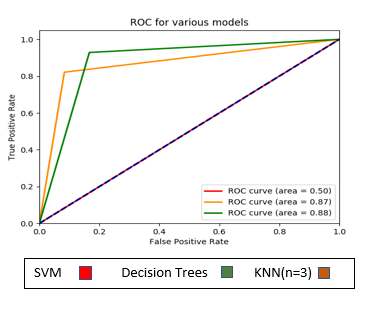}
  \caption{ROC Curve}}
\end{figure}
\newline
\subsubsection{\textbf{Finding zero crossings in the first derivative of each segment}}
\begin{itemize}
\item The path taken by the rider is divided into segments. The length of each segment is 1 percent of the total distance taken by the rider.
\item In each of these segments, the first derivative is found.
\item The points of zero crossings in the first derivative is identified. The number of points of zero crossings will be more for a squiggly path compared to a straight path due to presence of greater number of local maxima/minima in the segments. Figure 9 and 10 show the number of points of zero crossings for segments in a squiggly and a straight path. It can be seen from these figures, that segments in a straight path has zero or only 2 points of zero crossings, whereas segments in a squiggly path has 5 or more points of zero crossings.
\begin{figure}
  \textbf{\includegraphics[width=\linewidth]{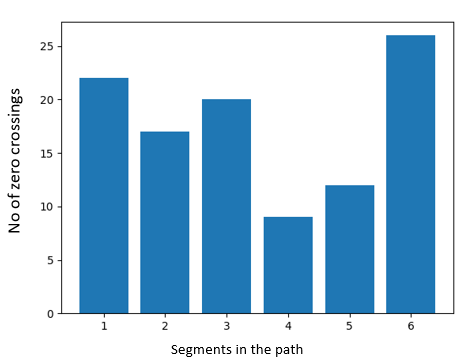}
  \caption{Number of Zero Crossings in segments of a squiggly path}}
\end{figure}
\begin{figure}
  \textbf{\includegraphics[width=\linewidth]{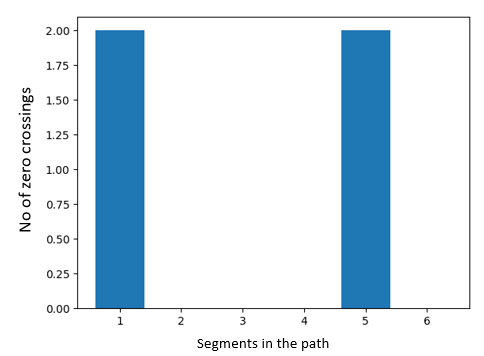}
  \caption{Number of Zero Crossings in segments of a straight path}}
\end{figure}
\end{itemize}
For each of the three models, the Table 1 shows the summary statistics when the ratio of train to test is 55:45
\begin{table}
\begin{center}
 \begin{tabular}{||c c c c||} 
 \hline
 Model & Accuracy & Precision & Recall\\ [0.1ex] 
 \hline\hline
 SVM &   62.74\% & 63\% & 63\%\\ 
 \hline
 KNN with K=3 &  56.73\% & 57\% & 57\%\\
 \hline
 Decision Trees & 71\% & 74\% & 71\%\\
 \hline
\end{tabular}
\textbf{\caption{Classification Report for method 3}}
\end{center}
\end{table}
From ROC curve analysis seen in Fig 11 it can be observed that Decision trees performed better than other classification models.
\begin{figure}
  \textbf{\includegraphics[width=\linewidth]{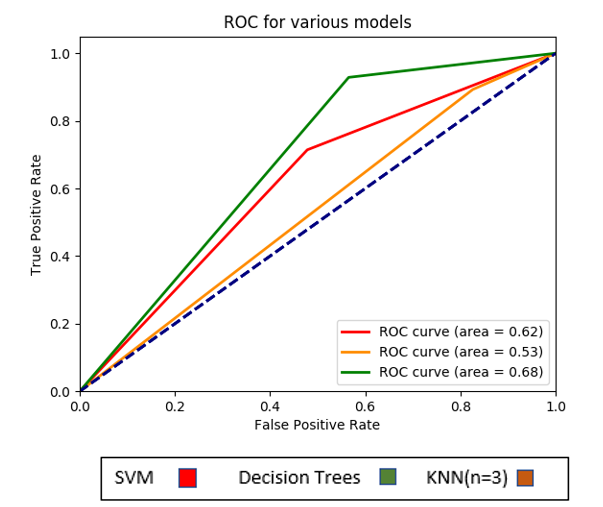}
  \caption{ROC Curve}}
\end{figure}
\newline
The Machine Learning models used for the above estimation are implemented using sklearn.svm.SVC \cite{c8} for Support vector machines, sklearn.neighbors.KNeighborsClassifier \cite{c9} for K nearest neighbors, sklearn.tree.DecisionTreeClassifier \cite{c10} for Decision Trees and sklearn.linear\_model.LinearRegression \cite{c11} for linear regression model.
\section{Results}
As seen from the above ROC plots, among the three methods, decision trees performed better and has more area under the ROC curve when the method of fitting a linear regression model for each segment was considered.
The above method was used to classify segments of the ride which was more squiggly than the other segments of the road.
 The Figure 12 shows segments marked in red which is more squiggly, the blue segments in the figure shows the part of the road which is straight.
\begin{figure}
  \textbf{\includegraphics[width=\linewidth]{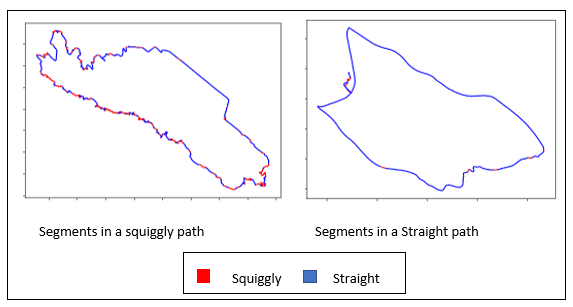}
  \caption{Segmented path}}
\end{figure}

The above information can also  be used to profile the user based on whether he prefers to ride on a dirt path or a bike road. In Figure 13 we show the percentage of distribution of squiggly(dirt) roads and straight roads taken by a specific athlete who donated data from Strava.

\begin{figure}
  \textbf{\includegraphics[width=\linewidth]{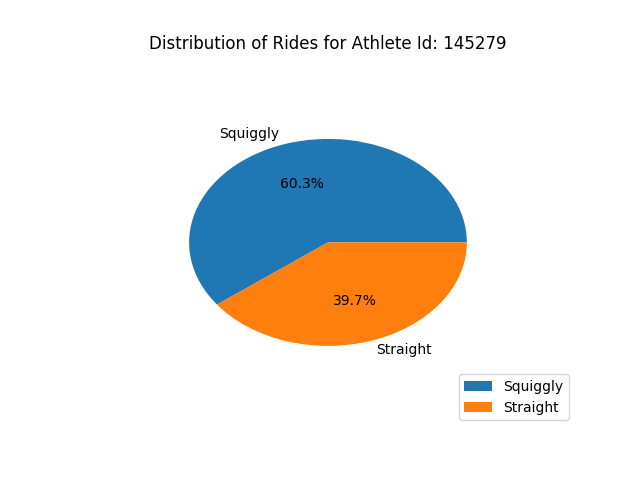}
  \caption{Distribution of type of path during rides.}}
\end{figure}

\section{CONCLUSIONS AND FUTURE WORK}
The paper illustrates the use of a simple method to estimate the type of road surface. The data used for the above experimentation can be obtained in a very easy and economical way. The type of surface of the road can be  used in many applications. One such includes the estimation of coefficient of friction and rolling resistance. Determining the coefficient of rolling resistance can be used in the estimation of Power against rolling resistance which can be used in total power estimation of the athlete during the ride as given by equation 1.
\newline 
	\textbf{\textit{P(total) = P(rolling resistance) + P(wind) +(gravity) + \\ P(acceleration)}\cite{c7} ...................................................... Eqn(1)} \newline
    
    Estimating the power output of an athlete can be used to estimate the VO2 of the individual as indicated by the below formula: \newline
    
    \textbf{\textit{vo2(max) = (10.8 * power/weight) + 7}................. Eqn(2)} \newline

    Commonly, VO2 estimation is done using the data from various sensors on fitness tracking wearable devices. Heart rate is one of the important attributes to determine VO2 max. A neural network can be built for estimating the VO2, by considering the heart rate variability as done in \cite{c12}. Accelerometer readings can also be combined with heart rate readings for VO2 estimation\cite{c13}. However, these wearables are expensive and not every user may own it. It is also inconvenient for the user to use the wearables continuously. In the future, we aim to use data that is freely available to us, such as wind speed, humidity, oxygen level and coefficient of friction from the type of surface to see how the power output can be modeled with respect to the above attributes. By estimating power output, we can estimate the VO2 max of the athlete during that ride as given in equation(2).\newline
        
    Another avenue of future work is in building recommendation systems for the users to suggest activities. Because certain cyclists may prefer smooth roads, versus mountain bike riders who prefer dirt roads, we can being to understand user preferences for bicycle travel paths. Other relevant insights can be drawn from the data we collect through Strava, weather, and social media platforms. Hence at the end of analysis, we will know the weather preference of the user, the kind of roads he prefers to cycle on, the time during which he regularly performs the activity and much more. These can be combined in order to recommend suitable activities to improve the user's health condition.\newline

\addtolength{\textheight}{-12cm}   




\end{document}